\documentclass[letterpaper, 10 pt, conference]{ieeeconf}  

\IEEEoverridecommandlockouts                              

\usepackage{amsmath}
\usepackage{amssymb}
\usepackage{graphicx}
\usepackage{float}
\usepackage{array}
\usepackage{tikz}
\usepackage{listings}
\usepackage{mathrsfs}
\usepackage{pgfplots}
\usepackage{python}
\usepackage[pdfencoding=auto,psdextra,pagebackref,breaklinks,colorlinks]{hyperref}
\usepackage{graphicx}
\usepackage{subcaption}
\usepackage{cite}
\usepackage{dsfont}
\usepackage{mathtools,etoolbox}
\usepackage{algorithm}
\usepackage{algpseudocode}

\usepackage[none]{hyphenat}
\usepackage[utf8]{inputenc}
\usepackage[english]{babel}

\usepackage{amsthm}
\usepackage{xfrac}
\usepackage{nicefrac}
\usepackage{xcolor}
\usepackage{booktabs}
\usepackage[switch, pagewise]{lineno}
\usepackage{nicefrac}
\usepackage{flushend}
\usepackage{gensymb}
\usepackage{lipsum}
\usepackage{balance}

\usepackage{pgfplots}
\pgfplotsset{compat=1.18}

\DeclareUnicodeCharacter{2212}{-}
\usepackage{caption} 

\title{\LARGE \bf
ViewActive: Active viewpoint optimization from a single image}

\author{Jiayi Wu, Xiaomin Lin, Botao He, Cornelia Fermüller, Yiannis Aloimonos %
\thanks{This work was supported by USDA NIFA sustainable agriculture system program under award number 20206801231805.}%
\thanks{Perception and Robotics Group, University of Maryland Institute for Advanced Computer Studies, University of Maryland, College Park, MD 20742, USA. 
Emails: \texttt{\{jiayiwu, xlin01, botao, fermulcm, jyaloimo\}@umd.edu}.
}
}

\begin{document}
\makeatletter
\g@addto@macro\@maketitle{
\begin{figure}[H]
  \setlength{\linewidth}{\textwidth}
  \setlength{\hsize}{\textwidth}
    \centering
    \includegraphics[width=0.95\textwidth]{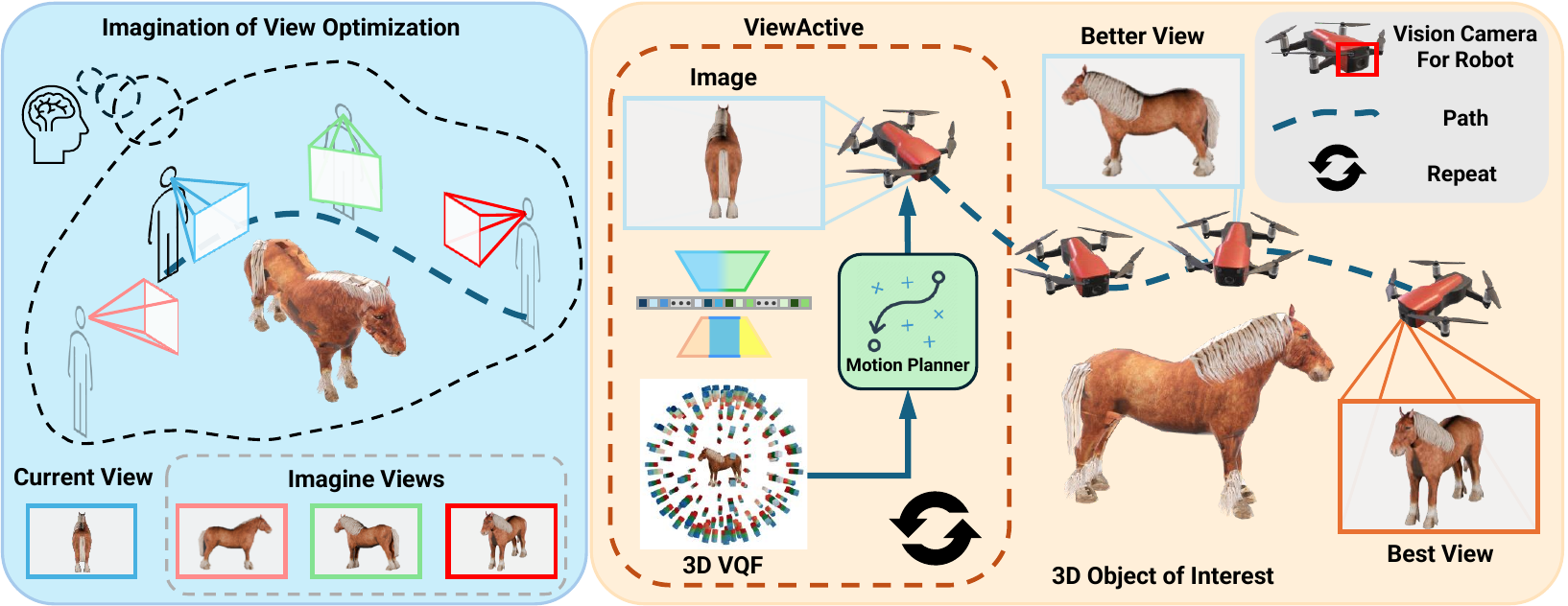}
    \captionsetup{font={footnotesize},labelfont=bf}
    \caption{The left side illustrates a humans’ ability of spatial visualization, where the system predicts multiple potential improved viewpoints from current 2D observation and selects the most informative one to update. To equip robots with this human-like capability for active viewpoint optimization, we propose the \textbf{ViewActive} pipeline, as shown on the right. The pipeline starts with an initial 2D image and predicts the 3D Viewpoint Quality Field (VQF) of the object of interest. Using latest prediction at each step, the motion planner progressively adjusts the robot's position to improve the visibility of the object.}
    \vspace{-5mm}
    \label{fig:Banner}
    \end{figure}
}
\maketitle

\addtocounter{figure}{-1} 
\thispagestyle{empty}
\pagestyle{empty}


\begin{abstract}
When observing objects, humans benefit from their spatial visualization and mental rotation ability to envision potential optimal viewpoints based on the current observation. 
This capability is crucial for enabling robots to achieve efficient and robust scene perception during operation, as optimal viewpoints provide essential and informative features for accurately representing scenes in 2D images, thereby enhancing downstream tasks.

To endow robots with this human-like active viewpoint optimization capability, we propose ViewActive, a modernized machine learning approach drawing inspiration from aspect graph, which provides viewpoint optimization guidance based solely on the current 2D image input.
Specifically, we introduce the 3D Viewpoint Quality Field (VQF), a compact and consistent representation for viewpoint quality distribution similar to an aspect graph, composed of three general-purpose viewpoint quality metrics: self-occlusion ratio, occupancy-aware surface normal entropy, and visual entropy.
We utilize pre-trained image encoders to extract robust visual and semantic features, which are then decoded into the 3D VQF, allowing our model to generalize effectively across diverse objects, including unseen categories.
The lightweight ViewActive network ($72$ FPS on a single GPU) significantly enhances the performance of state-of-the-art object recognition pipelines and can be integrated into real-time motion planning for robotic applications. Our code and dataset are available here \url{https://github.com/jiayi-wu-umd/ViewActive}.


\end{abstract}
\section{INTRODUCTION}

Humans can anticipate the best viewpoints for a given object from only a single observation, thanks to their extensive experience and observation of various objects. This benefits from humans' ability of spatial visualization and mental rotation \cite{tarr1989mental,wexler1998motor} (see the left half of Fig. \ref{fig:Banner}), which allows them to proactively adjust their position to gather more information. The viewpoint's quality significantly impacts a robot's perception and interaction with the surroundings. Poorly chosen viewpoints, often referred to as ``accidental views"\cite{gigus1991efficiently, plantinga1990visibility}, can obscure critical features of an object. In contrast, ``canonical views" \cite{palmer1981cannonical} provide perspectives that reveal the most informative features of an object, significantly enhancing the robot's performance in tasks such as object detection\cite{wang2024yolov10,zou2023object,wu2024marvis}, 
navigation\cite{lin2024uivnav,wu2025single}, 3D reconstruction \cite{xiongevent3dgs,wu20233d}, and manipulation\cite{billard2019trends,siddique2025aquafuse}. Therefore, the ability to actively avoid accidental viewpoints and obtain better viewpoint is crucial for robots. This is known as active perception\cite{bajcsy2018revisiting,aloimonos2013active}, \textit{``where a robot purposefully selects what to perceive, how, and when to perceive it, driven by its current goals and interaction with the environment"}.

However, modeling this 
capability into an algorithm that a robot can execute is highly challenging. First, the camera can only provide limited visual information at the current moment (a single 2D image). Second, the criteria humans use to evaluate the quality of a viewpoint are often vague and may vary depending on the task. Translating these vague criteria into a clear and task-agnostic quantitative metric is essential to successfully identify the optimal viewpoint.

In this study, we demonstrate that robots, similar to humans, can actively optimize the viewpoint by embedding viewpoint quality distribution priors into the network. In our approach, \textbf{ViewActive}, an active vision system predicts a ``better" viewpoint of the target object without physical movement. We design a supervised machine learning approach that enables robots to estimate the 3D distribution of viewpoint quality of various objects (see the right half of Fig. \ref{fig:Banner}). Based on a single camera observation, our method efficiently predicts the 3D Viewpoint Quality Field (VQF). During the optimization, our method accounts for the robot's limitations (lack of global information and restricted movement) while leveraging predicted 3D VQF to progressively optimize viewpoints, enhancing downstream tasks. ViewActive is reminiscent of the aspect graph approach that dominated object recognition research in the 1990s \cite{gigus1991efficiently,plantinga1990visibility} but is now framed in the machine learning paradigm. Conceptually, ViewActive mirrors the process of constructing an aspect graph, as if it imagines the aspect graph and selects the next best viewpoint based on a single image. The contributions of this paper are as follows:

\begin{itemize}
\item Three general-purpose viewpoint quality metrics to comprehensively and robustly quantify viewpoint quality: self-occlusion ratio, occupancy-aware surface normal entropy, and visual entropy. 
\item A learnable representation to structurally and consistently represent the 3D viewpoint quality distribution of objects: 3D Viewpoint Quality Field (VQF).
\item A lightweight viewpoint optimization pipeline based on a single image of each step, consisting of a generalizable and class-agnostic 3D VQF estimator and a reachable-aware progressive viewpoint optimizer. 
\end{itemize}

\section{Background \& related works}
The concept of ``viewpoint" has been extensively discussed in computer graphics and virtual reality. The ability to select an optimal viewpoint is essential for accurately perceiving and interacting with objects in downstream tasks \cite{yuan2024learning,yu2022udepth,wu2023low}. In this section, we begin with the methods used to evaluate viewpoint quality and then discuss the challenges of selecting and optimizing viewpoints.

\label{section:related_work}
\subsection{Viewpoint Quality Evaluation}
\begin{figure*}[h]
\vspace{3mm}
\includegraphics[width=0.9\textwidth]{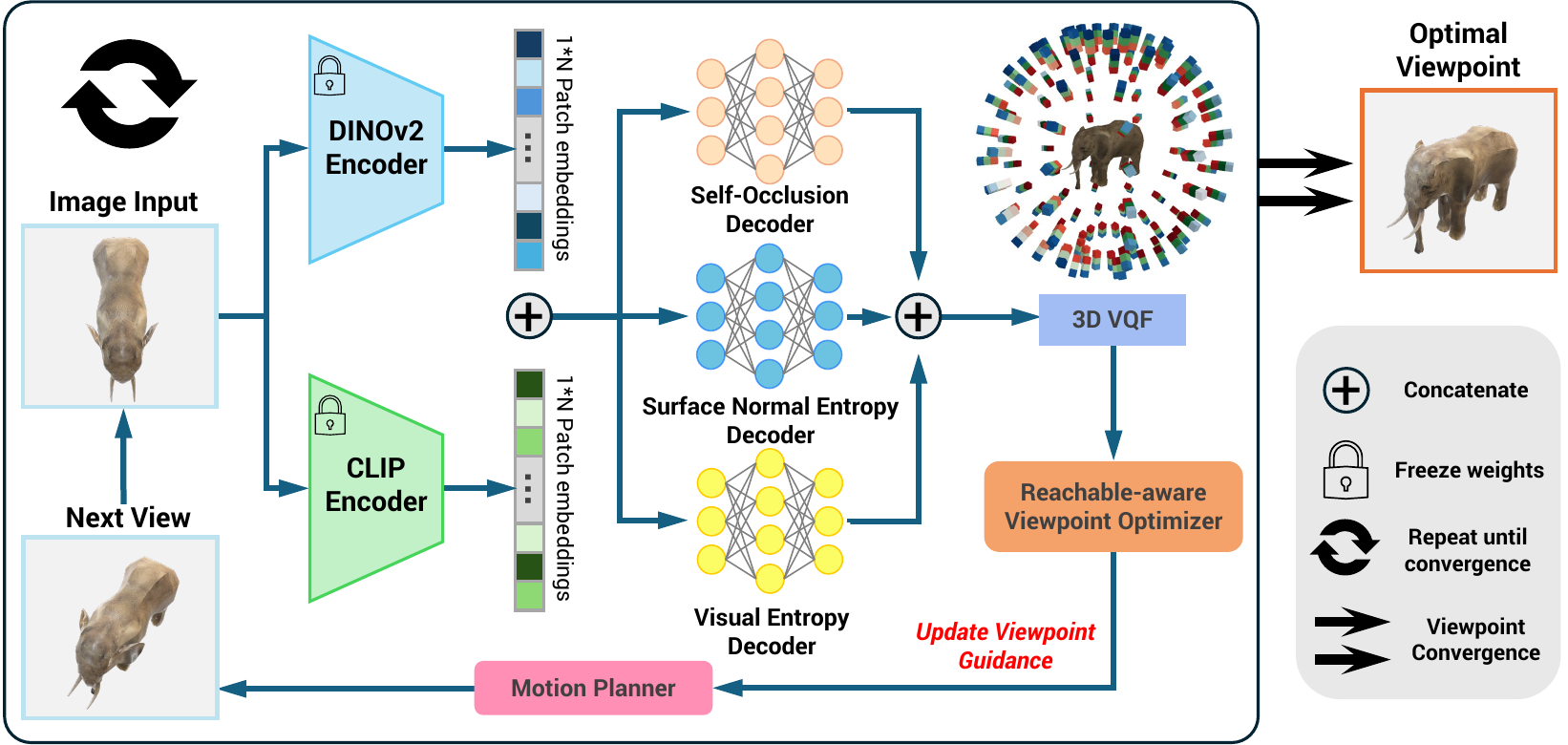}
\centering
\captionsetup{font={footnotesize},labelfont=bf}
\caption{Architecture of the ViewActive learning pipeline. The pipeline takes an image as input and leverages DINOv2\cite{oquab2023dinov2} and CLIP\cite{radford2021learning} encoders to extract robust visual and semantic features. These features are processed by three lightweight decoders to estimate the 3D Viewpoint Quality Field (VQF) for object observation. The viewpoint optimizer performs a single optimization step and sends the result to the motion planner for movement. The network continuously refines the viewpoint guidance based on current observation until it converges on the optimal view of the object.}
\label{fig:ViewActive_arch}
\vspace{-5mm}
\end{figure*}
Currently, we do not have a unified definition of a "good viewpoint" \cite{zhang2019overview,roldao20223d}. Since Palmer et al. first introduced the concept of a "Canonical Perspective" \cite{palmer1981canonical} in 1981, a wealth of work has emerged on how to evaluate viewpoints and select the best viewpoint. Various existing viewpoint theories have evaluated viewpoints based on different criteria, such as area \cite{Plemenos1996, vazquez2001viewpoint}, silhouette \cite{vieira2009learning, feldman2005information}, depth \cite{stoev2002case}, surface curvature \cite{page2003shape,leifman2016surface}, semantics \cite{laga2010semantics,attene2006mesh}, aesthetics\cite{zhang20203d}, etc.

With the advancement of active vision, recent work has begun to explore methods for improving the performance of specific visual tasks through active viewpoint adjustments. However, the viewpoint quality metrics used in these approaches are often tailored to particular downstream tasks, such as object grasping \cite{gualtieri2017viewpoint}, human pose estimation \cite{Kiciroglu_2020_CVPR,chen2024active}, etc. 
Task-specific metrics, such as output precision, consistency, and uncertainty, are effective within their contexts but difficult to generalize across tasks. This raises the question of whether viewpoint quality should be defined relative to downstream task performance or evaluated by broader, general-purpose metrics.

To address this, we redesign three task-agnostic viewpoint quality metrics to comprehensively and uniformly evaluate viewpoints. These metrics not only exhibit high consistency across a wide variety of objects, but are also compatible with the capabilities of current mainstream simulators and common machine learning paradigms. 

\subsection{Viewpoint Selection and Optimization}
Choosing the optimal viewpoint has been a fundamental challenge in the field of computer vision and robotics.
In early research, many works \cite{vazquez2001viewpoint,leifman2016surface} focused on selecting the optimal viewpoint from multiple predefined viewpoints, assuming that the 3D model of the object is known. Song\cite{song2022unsupervised} proposed a new unsupervised view selection algorithm that addresses the challenge of selecting salient views and detecting 3D interest points. Zhang\cite{zhang20203d} focused on optimizing the viewpoints to achieve high aesthetic quality in rendered 3D images. These methods are suitable for scenarios where the environment is known and only offline inference is required, such as in computer games and virtual reality. However, this assumption is often difficult to meet in robotics, where global knowledge is not given (only the image from the current viewpoint is available) and online inference is needed.  

In recent years, some active vision methods for viewpoint optimization tailored to specific visual tasks have been proposed. Kiciroglu et al. \cite{Kiciroglu_2020_CVPR} introduced a method for selecting the next best viewpoint based on a sequence of continuous 2D images by estimating and minimizing uncertainty in the estimation of 3D body pose. Chen et al.~\cite{chen2024active} improved upon this by directly selecting the best viewpoint to minimize error in 2D human pose estimation, eliminating the need for 3D pose estimation. He et al. \cite{siming2024active} introduced a method using Neural Radiance Fields (NeRF) \cite{mildenhall2020nerfrepresentingscenesneural} to synthesize views and maximize predictive information for exploration in cluttered environments. While effective, these methods are limited to specific tasks and lack generalization across diverse objects.

In this work, we introduce an active viewpoint optimization framework that utilizes only a single 2D image, namely ViewActive.
By leveraging the robust visual and semantic priors from pre-trained image encoders and the compact viewpoint quality representation from our proposed 3D Viewpoint Quality Field (VQF), ViewActive achieves effective active viewpoint optimization that generalizes across diverse object categories, even unseen ones.


\section{Methodology}

Having established an understanding of viewpoint quality, we first introduce general-purpose viewpoint quality metrics. These metrics are used to construct the 3D Viewpoint Quality Field (VQF). Next, we describe our data collection process and detail the network design for estimating the 3D VQF. Finally, we incorporate reachable-aware progressive optimization in our viewpoint optimization process.
\label{section:methodology}

\subsection{General-purpose Viewpoint Quality Metrics} \label{General-purpose Viewpoint Quality Metrics}
To comprehensively and robustly quantify viewpoint quality in the general sense, we propose three task-agnostic metrics that include 3D geometric attributes and 2D visual features of observed objects from various viewpoints. 
\subsubsection{\textbf{Self-occlusion Ratio}}
Occlusion is a critical factor affecting viewpoint quality \cite{Blanz1999WhatOA}. For an object, a lower self-occlusion ratio (i.e., a higher proportion of visible surface area relative to the total surface area) generally reveals richer visual and structural characteristics, making the object easier to distinguish. Based on this observation, we use the objects' self-occlusion ratio from a given viewpoint as a general-purpose viewpoint quality metric, defined as follows:
\begin{equation}\label{eq:1_Self_occlusion}
\mathcal{R}_{occlu}(v) = \frac{\sum{A_{Occluded}(v)}}{\sum{A_{All}(v)}},
\end{equation}, where $\mathcal{R}_{occlu}(v)$ represents the self-occlusion ratio of the target object from the viewpoint $v$, $\sum{A_{Occluded}(v)}$ is the total occluded surface area of the object from viewpoint $v$, and $\sum{A_{All}(v)}$ is the total surface area of the object. A smaller value of $\mathcal{R}_{occlu}(v)$ indicates that a larger portion of the object is visible, providing more information about the object’s geometry and appearance. While this metric effectively measures visibility, surface area alone does not always convey sufficient structural complexity (e.g., flat surfaces). To address this, we introduce a complementary metric focused on geometric richness.

\subsubsection{\textbf{Occupancy-aware Surface Normal Entropy}}
Upon further analysis, a higher proportion of visible surface area rarely indicates richer geometric structural features. Therefore, to measure geometric structural richness carried by the visible surface area from different viewpoints, we designed another metric: \textbf{Occupancy-aware Surface Normal Entropy} (inspired by the concept of information entropy) $H_{geo}$. 

For each viewpoint, we calculate the occupancy ratio and entropy of the 3D normals of all visible surfaces (polygons) within the visible range. The visible range refers to the range of directions in which the normal vectors of the visible surfaces are distributed, originating from the object's center and extending in the hemisphere towards the camera's direction. Through extensive experimentation, we find that this metric consistently represents the amount of geometric information on the visible surface and is highly correlated with viewpoint quality. The detailed mathematical expressions can be written as follows:
\begin{equation}\label{eq:2_Normal_Entropy}
H_{geo}(\mathbf{\Theta,\Phi}|v) = - \frac{N_{occup}(v)}{N_{all}} \cdot \sum_{\theta,\phi \in \mathbf{\Theta,\Phi}}p(\theta,\phi)\log_{2} p(\theta,\phi),
\end{equation}
where $\mathbf{\Theta}$ and $\mathbf{\Phi}$ are, respectively, the patch of normal vectors' polar angles and azimuthal angles in spherical coordinate system. $N_{occup}(v)$ denotes the number of patches occupied by the normal vectors of visible surfaces from the viewpoint $v$ and $N_{all}$ represents the total number of patches within the visible range, $\theta$ and $\phi$ denote the possible values of the normalized normal vector's polar angles and azimuthal angles, $\mathbf{\Theta}\rightarrow[0,\pi/2]$, $\mathbf{\Phi}\rightarrow[-\pi,\pi]$. The probability $p(\theta,\phi)$ is calculated from the normalized 2d histogram counts of the variables $\theta$ and $\phi$.

\subsubsection{\textbf{Visual Entropy}}
In addition to geometric features, the amount of information carried by visual features on the visible surfaces (such as grayscale intensity, color, and texture) is also a significant factor influencing viewpoint quality. Most existing robotic vision primary inputs are still 2D image sequences, with feature extraction typically performed in pixel space (RGB, grayscale, etc.). Extensive experiments reveal that viewpoints with rich visual features on visible surfaces enhance object recognition.  Based on these observations and analyses, we formulate the third viewpoint quality metric, \textbf{Visual Entropy} $H_{vis}$, to quantify the information in visual features under different viewpoints.
\begin{equation}\label{eq:3_Visual_Entropy}
H_{vis}(\mathbf{G}|v) = - \sum_{g \in \mathbf{G}}p(g)\log_{2} p(g),
\end{equation}
where $\mathbf{G}$ is the patch of greyscale levels, $g$ denotes the possible values of the greyscale, $\mathbf{G}\rightarrow[0,1]$. The probability $p(g)$ is calculated from the normalized histogram counts of the variables $g$.

\begin{figure}[t]
\vspace{3mm}
\includegraphics[width=0.9\linewidth]{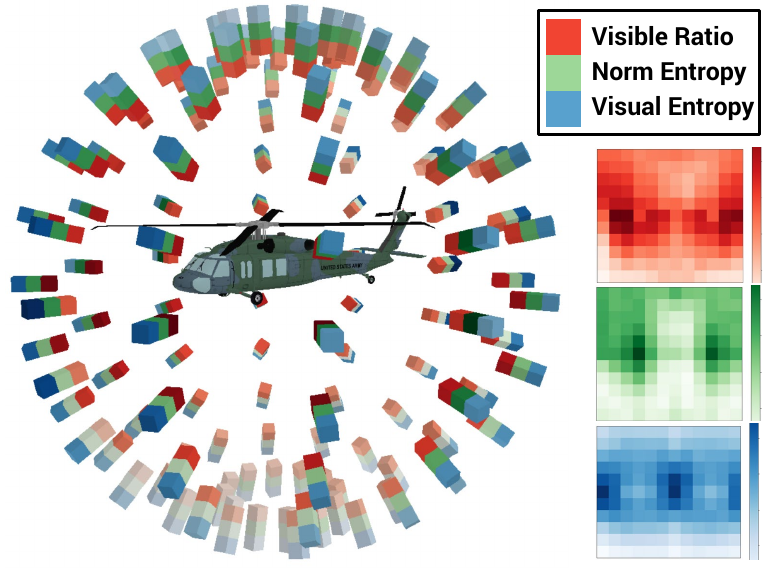}
\centering
\captionsetup{font={footnotesize},labelfont=bf}
\caption{3D Viewpoint Quality Field (VQF). The 3D VQF represents the distribution of viewpoint quality around an object, in this case, a helicoptor, within a spherical camera space. Each $1 \times 3$ colored block corresponds to the quality of a specific viewpoint representing all three general-purpose viewpoint quality metrics. This structured field helps guide the system in selecting the most informative viewpoints for observation and analysis. The heatmaps on the right-hand side shows metrics values for all 132 views.  \textit{Hint: Visible Ratio is $(1- \mathcal{R}_{occlu})$}.}
\label{fig:3D_VQF}
\vspace{-5mm}
\end{figure}

\subsection{3D Viewpoint Quality Field (VQF)}
We utilize the metrics introduced in Sec.~\ref{General-purpose Viewpoint Quality Metrics} to construct the 3D Viewpoint Quality Field, which structurally represents the distribution of viewpoint quality for objects. As shown in Fig.~\ref{fig:3D_VQF}, within a spherical camera viewpoint space, the 3D VQF encodes the general-purpose viewpoint quality distribution for all predefined candidate camera viewpoints. The quality of each candidate viewpoint is characterized by a fixed-length ($1 \times n$) viewpoint quality feature vector, which is later used to train our \textbf{ViewActive} network.

\subsection{Data Collection} 
To train a robust network to predict the 3D VQF, we automatically collected data using 3D models from the Objaverse dataset \cite{deitke2022objaverseuniverseannotated3d} with LVIS annotations. Although Objaverse contains millions of models, only 46,000 have class annotations. For this study, we selected 1,600 models from various classes represented in the ImageNet dataset \cite{deng2009imagenet}.
Each model was loaded into Blender$^{\text{TM}}$\cite{blender} and standardized by sanitizing, scaling, and centering. We employ a spherical viewpoint sampling strategy along with ray-casting, capturing the corresponding viewpoint quality feature vector of 132 views per object by varying azimuthal and polar angles. Two images are rendered for each view: an RGB image of the object and a mask to isolate it from the background.

\vspace{-1mm}
\subsection{Class-agnostic 3D VQF Estimation Network}\label{Class-agnostic 3D VQF Estimation from A Single 2D Observation}

The robust viewpoint quality metrics we proposed and the automated data generation pipeline make it possible to consistently capture 3D VQF across a wide variety of objects on a large scale in simulation. Building on this, we designed a lightweight network (see Fig.~\ref{fig:ViewActive_arch}) that learns from the data to robustly estimate the 3D VQF based solely on a single image. For the encoders, we utilize both the pre-trained DINOv2 \cite{oquab2023dinov2} and CLIP \cite{radford2021learning} image encoders to provide our pipeline with robust image feature embeddings in the latent space. Specifically, DINOv2 provides the pipeline with general-purpose visual features by capturing a wide range of visual patterns and structures through its self-supervised learning approach. Meanwhile, CLIP contributes general semantic features by leveraging contrastive learning to align visual inputs with textual descriptions, thereby improving the semantic understanding of images. We integrate these visual and semantic features within our network to serve as the latent representation of the input image. This robust representation contains rich underlying information and enables our network to exhibit strong generalization capabilities and zero-shot inference across diverse unseen image domains and object categories. 

To avoid interference, we employ a multi-stream decoder architecture to independently learn each channel of the 3D VQF. Each decoder consists of a 3-layer MLP ($0.5M$ parameters), allowing the network to specialize in learning different components of viewpoint quality features while maintaining a lightweight design. Since the relative values of viewpoint quality features guide optimization, while absolute values are less important, we design a relaxed loss function to help the network learn the required capability. 
The loss function consists of the following three components: 
\begin{itemize}
\item Mean absolute error: $\mathcal{L}_{1} =  \Vert \widetilde{\mathbf{q}} - \mathbf{q} \Vert_1$
\item Structural dissimilarity: $\mathcal{L}_{DSSIM} = DSSIM(\widetilde{\mathbf{q}}, \mathbf{q})$
\item Scale-invariant log: $\mathcal{L}_{SILog} = \frac{1}{N}\sum_i[\log(\widetilde{q_i})-\log(q)]^{2} - \frac{\lambda}{N^2}[\sum_{i}\log(\widetilde{q_i})-\log(q_i)]^{2}$,
\end{itemize}
where $\widetilde{q}$ is the predicted viewpoint quality feature value, $q$ is the ground truth of the viewpoint quality feature value, $N$ denotes the number of valid viewpoints, and we set we $\lambda = 0.85$ in our implementation. Finally, the end-to-end learning objective of our network is formulated as:
\begin{equation}\label{eq4:total_loss}
\mathcal{L}_{total} = \lambda_{1}\mathcal{L}_{1} + \lambda_{2}\mathcal{L}_{DSSIM} + \lambda_{3}\mathcal{L}_{SILog}.
\end{equation}
we find the $\lambda$-parameters empirically through
hyper-parameter tuning. The optimal values used in
ViewActive training are $\lambda_{1} = 0.3$, $\lambda_{2} = 0.4$, and $\lambda_{3} = 0.3$. 

\subsection{Reachable-aware Progressive Viewpoint Optimization}

The proposed 3D VQF estimation network in Sec.~\ref{Class-agnostic 3D VQF Estimation from A Single 2D Observation} endows the robot with a global viewpoint quality estimation capability based on a single image, guiding the next-step viewpoint optimization. Although our method can estimate the optimal viewpoint from a single image, considering the robot's movement speed limitations and the potential dynamism of objects in the scene, we incorporate reachability constraints into the viewpoint optimization. This adjustment allows our viewpoint optimization pipeline to perform online inference in dynamic environments. Moreover, for static objects, improved viewpoints also lead to better 3D VQF estimation. Progressive optimization ensures consistent improvement in both the 3D VQF estimation pipeline's performance and the downstream task. The details of the optimization algorithm are shown in Algorithm \ref{alg:viewpoint_opt}.
\vspace{-2mm}
\begin{algorithm}[h]
\caption{Reachable-aware Viewpoint Optimization}\label{alg:viewpoint_opt}
\begin{algorithmic}
\Require $v_{c} = (\theta_{c},\phi_{c})$ - Current Viewpoint, 
        \newline $\mathbf{M}_{VQF}$ - Estimated VQF,
        \newline $[\mathbf{\theta_{r}},\mathbf{\phi_{r}}]$ - A Set of Reachable Viewpoints,
\Ensure $v_{n} = (\theta_{n},\phi_{n})$ - Next Viewpoint
\While{$v_{n} \neq v_{c}$}
\If{$\arg \max_{\theta, \phi}(\mathbf{M}_{VQF}) \in [\mathbf{\theta_{r}},\mathbf{\phi_{r}}] $}
    \State $v_{n} \gets \arg \max_{\theta, \phi}(\mathbf{M}_{VQF})$
\Else
    \State $v_{n} \gets \text{nearest $v$} \in [\mathbf{\theta_{r}},\mathbf{\phi_{r}}] \text{ }\text{to } \arg \max_{\theta, \phi}(\mathbf{M}_{VQF})$
\EndIf
\EndWhile
\end{algorithmic}
\end{algorithm}
\vspace{-2mm}

\section{Experiments}
\label{section:Experiments_and_results}
After the detailed explanation of ViewActive, we now move to the Experimental Section, where these concepts are rigorously evaluated qualitatively, quantitatively, and through a drone motion planning experiment. We begin with an overview of the dataset and implementation details.
\vspace{-2mm}
\subsection{Dataset and Implementation Details}
To construct the 3D Viewpoint Quality Field (VQF), we selected the top $56$ classes, resulting in 1600 3D models from which $211,200$ images ($1920\times1680$ pixels with masks) were generated. Additionally, to test the generalization ability of ViewActive on unseen object categories, we created a test set of $200$ objects ($26,400$ images) for unseen classes. We conducted dense sampling of 10 points per face on each mesh, employing a ray-casting method to assess occlusion and record the normal vectors of fully visible faces for further analysis. The data was split into training, validation, and test sets with a $70:5:25$ ratio, and cross-validation was performed during training. We used AdamW \cite{loshchilov2019decoupledweightdecayregularization} as the optimizer with an initial learning rate of $0.001$, applying exponential decay with a multiplicative factor of $0.9$.

\begin{figure}[t]
\vspace{2mm}
\includegraphics[width=0.98\linewidth]{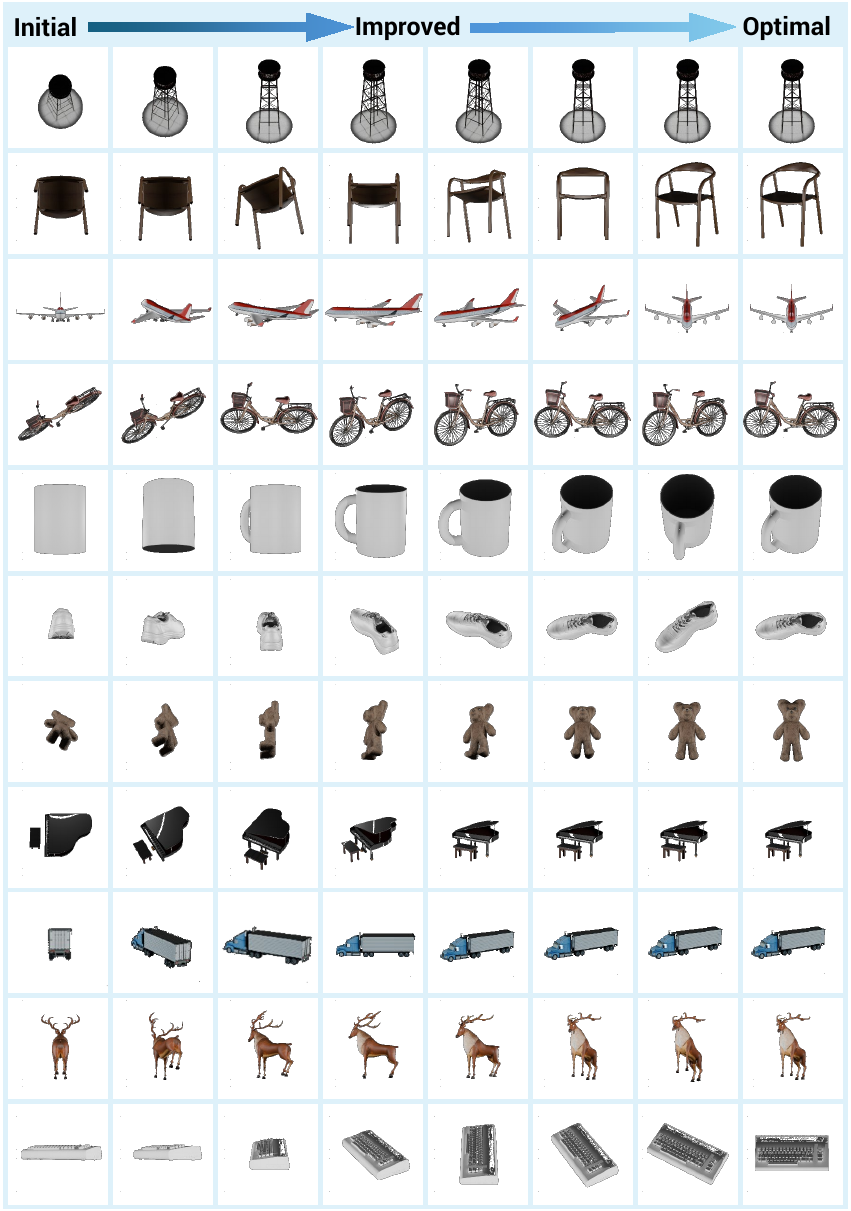}
\centering
\captionsetup{font={footnotesize},labelfont=bf}
\caption{Active Viewpoint Optimization, starting from Left to Right.}
\label{fig:qual_eval}
\vspace{-4mm}
\end{figure}
\addtocounter{figure}{+1} 

\begin{figure*}[h]
\vspace{2mm}
\includegraphics[width=0.93\linewidth]{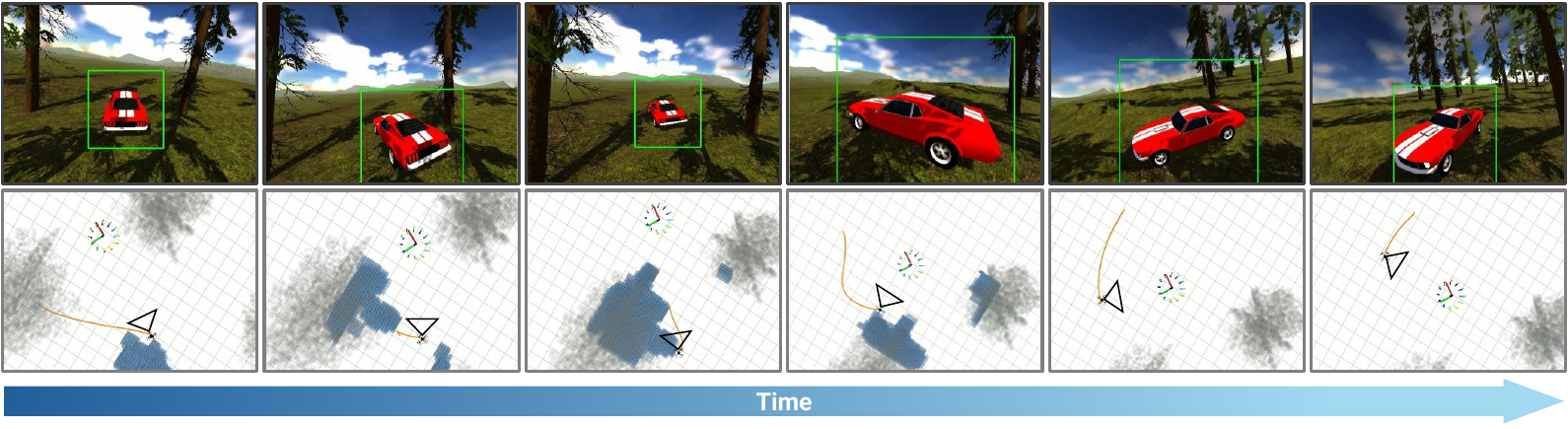}
\vspace{-2mm}
\centering
\captionsetup{font={footnotesize},labelfont=bf}
\caption{The drone followed an optimized path using active viewpoint selection and obstacle avoidance, ensuring safety and improved object visibility in a simulated environment with dynamic 3D VQF estimation.}
\label{fig:motion_planning}
\vspace{-5mm}
\end{figure*}
\addtocounter{figure}{-2} 
\begin{figure}[h]
\vspace{0mm}
\includegraphics[width=0.99\linewidth]{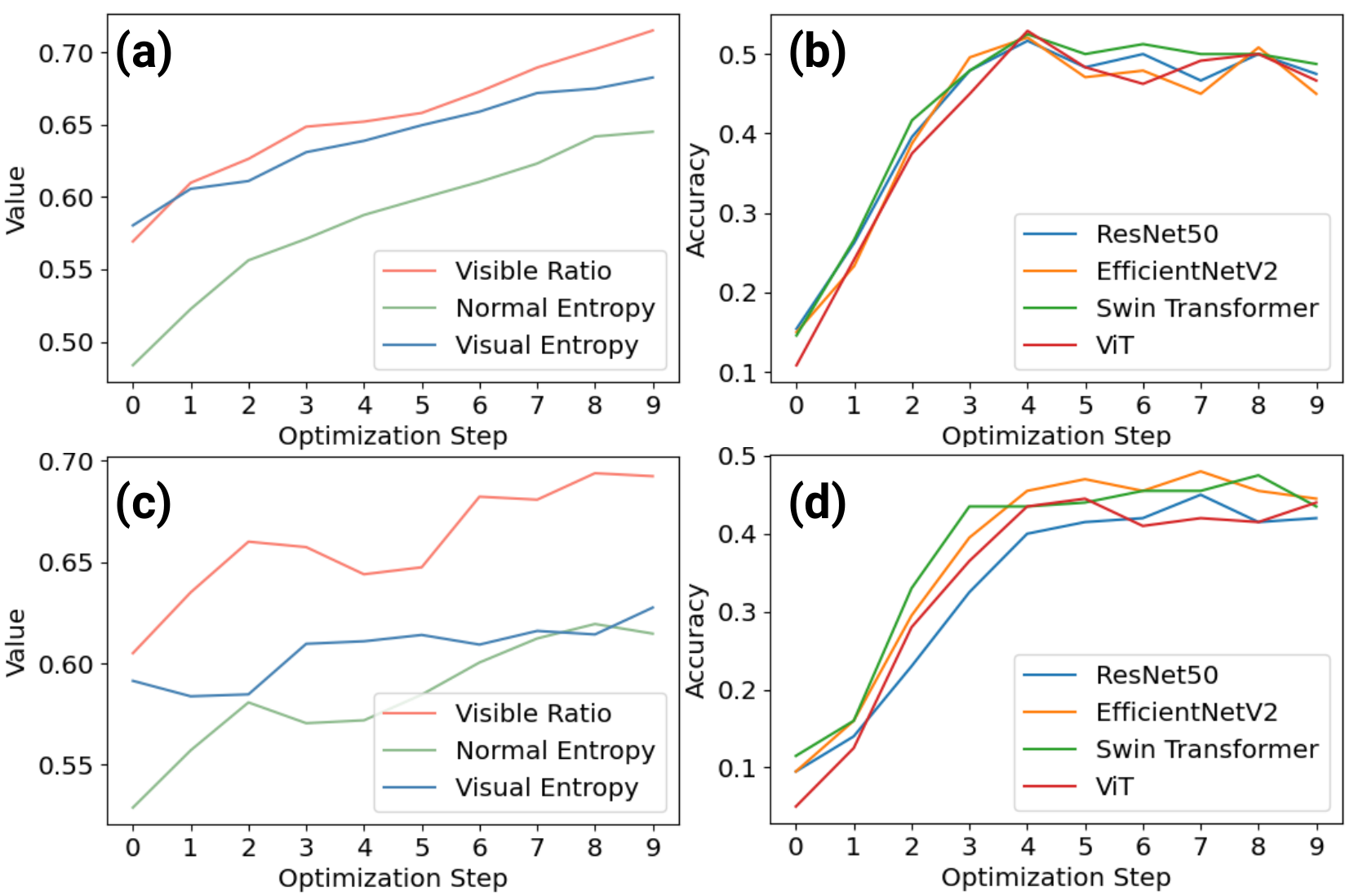}
\centering
\captionsetup{font={footnotesize},labelfont=bf}
\caption{\textbf{Quantitative results}: (a) Three viewpoint quality metrics' values for seen categories (b) Object recognition accuracy for seen categories. (c) Three viewpoint quality metrics' values for unseen categories. (d) Object recognition accuracy for unseen categories.}
\label{fig:quant_eval}
\vspace{-8mm}
\end{figure}

\subsection{Qualitative Evaluation}
To the best of our knowledge, no existing pipeline performs active viewpoint optimization based on a single image. Therefore, we qualitatively evaluated the performance of our viewpoint optimization across $7$ steps on a test set containing various objects. As shown in Fig.~\ref{fig:qual_eval}, 
the qualitative comparison of object viewpoints across $7$ steps showcases the gradual increase in the amount of information contained in the images from left to right. 

Initially, objects may be seen from accidental viewpoints with partial visibility or suboptimal perspectives. However, as ViewActive's progressive optimization progresses, the viewpoints gradually improve and converge on more robust and informative angles, making the critical features of each object increasingly distinguishable. The final column presents objects from a feature-rich and informative perspective, demonstrating ViewActive's effectiveness in accurate object perception and its ability to avoid accidental or obstructed views. 

\subsection{Quantitative Evaluation}

To demonstrate that our method can significantly improve performance on general object recognition tasks, we applied ViewActive as a viewpoint optimization pipeline across the entire test set, which includes both seen and unseen classes from the training set. We also selected four state-of-the-art (SOTA) object recognition networks for quantitative evaluation. For each sample, we randomly initialized the starting viewpoint and conducted $9$ rounds of iterative viewpoint optimization, running the object recognition pipeline at each step. As shown in Fig.~\ref{fig:quant_eval}\textcolor{red}{(a)} and \ref{fig:quant_eval}\textcolor{red}{(b)}, in the seen categories, both the accuracy of the object recognition models and the three evaluation metrics improved steadily as the viewpoints were progressively optimized. This confirms the correlation between the general-purpose viewpoint quality metrics we designed and object recognition task performance, and it also demonstrates the effectiveness of ViewActive in enhancing downstream task performance by predicting 3D VQF from a single image.

Additionally, to evaluate the generalizability of ViewActive on unseen categories (zero-shot inference), we constructed another evaluation on test set containing samples from categories not presented in the training set. As shown in Fig.~\ref{fig:quant_eval}\textcolor{red}{(c)} and \ref{fig:quant_eval}\textcolor{red}{(d)}, ViewActive provides reasonable viewpoint optimization guidance even for unseen categories. Furthermore, the three general-purpose evaluation metrics and the accuracy of the object recognition models show a clear upward trend, demonstrating the effectiveness and strong generalizability of our approach. Moreover, ViewActive can rapidly predict 3D VQF with an inference rate of $72$ FPS on a single NVIDIA$^{\text{TM}}$ RTX 4070 GPU.

\subsection{Drone Motion Planning Experiment}
We conducted a motion planning experiment within a simulated environment with numerous obstacles to assess the effectiveness of ViewActive in a robotic system and the advantages of its active viewpoint optimization capabilities. 
The environment and drone perception were simulated in Unity\cite{Unity2024}, while control and motion planning were simulated in ROS \cite{quigley2009ros} with Rviz \cite{kam2015rviz} for visualization. Specifically, the predicted 3D VQF was reprojected onto a sphere with a $5$m radius around the target object (see Fig.~\ref{fig:motion_planning}). 
At each step, the Euclidean distance between the drone and the viewpoint locations was added as a weight to the initial 3D VQF. The location with the highest value on the 3D VQF was designated as the next waypoint for the drone.  Since our motion planner avoids obstacles and minimizes objects obstructing the view of the target, we can guide the drone to better viewpoints while ensuring safety and visibility of the target object. As depicted in Fig.~\ref{fig:motion_planning}, the drone selects suboptimal viewpoints to avoid collisions. Once the obstacles are cleared, it navigates to a better viewpoint, providing richer visual features of the target object.

\section{CONCLUSIONS}
\label{section:Conclusions}
In this paper, we introduced \textbf{ViewActive}, a novel framework for active viewpoint optimization that mimics human-like spatial reasoning to enhance robotic perception. To achieve this, we developed the 3D Viewpoint Quality Field (VQF) by introducing three general-purpose viewpoint quality metrics—self-occlusion ratio, occupancy-aware surface normal entropy, and visual entropy. Our approach leverages pre-trained models to extract robust visual and semantic features from 2D images. Using these feature embeddings, we trained ViewActive to predict the 3D VQF, allowing robots to estimate the best viewpoints from a single 2D image. Through comprehensive experimental analysis, ViewActive significantly improves object recognition accuracy for both seen and unseen classes. This work demonstrates the potential of active perception techniques to optimize robotic scene understanding without the need for physical movement or complete 3D information.

Looking ahead, ViewActive can be extended to complex scenes involving multi-objects, and applied to real-world applications such as robotic manipulation and exploration for unknown environment.










\bibliographystyle{IEEEtran}
\bibliography{IEEEabrv,refs}

\begin{thebibliography}{10}
\providecommand{\url}[1]{#1}
\csname url@samestyle\endcsname
\providecommand{\newblock}{\relax}
\providecommand{\bibinfo}[2]{#2}
\providecommand{\BIBentrySTDinterwordspacing}{\spaceskip=0pt\relax}
\providecommand{\BIBentryALTinterwordstretchfactor}{4}
\providecommand{\BIBentryALTinterwordspacing}{\spaceskip=\fontdimen2\font plus
\BIBentryALTinterwordstretchfactor\fontdimen3\font minus \fontdimen4\font\relax}
\providecommand{\BIBforeignlanguage}[2]{{%
\expandafter\ifx\csname l@#1\endcsname\relax
\typeout{** WARNING: IEEEtran.bst: No hyphenation pattern has been}%
\typeout{** loaded for the language `#1'. Using the pattern for}%
\typeout{** the default language instead.}%
\else
\language=\csname l@#1\endcsname
\fi
#2}}
\providecommand{\BIBdecl}{\relax}
\BIBdecl

\bibitem{tarr1989mental}
M.~J. Tarr and S.~Pinker, ``Mental rotation and orientation-dependence in shape recognition,'' \emph{Cognitive psychology}, vol.~21, no.~2, pp. 233--282, 1989.

\bibitem{wexler1998motor}
M.~Wexler, S.~M. Kosslyn, and A.~Berthoz, ``Motor processes in mental rotation,'' \emph{Cognition}, vol.~68, no.~1, pp. 77--94, 1998.

\bibitem{gigus1991efficiently}
Z.~Gigus, J.~Canny, and R.~Seidel, ``Efficiently computing and representing aspect graphs of polyhedral objects,'' \emph{IEEE Transactions on Pattern Analysis and Machine Intelligence}, vol.~13, no.~6, pp. 542--551, 1991.

\bibitem{plantinga1990visibility}
H.~Plantinga and C.~R. Dyer, ``Visibility, occlusion, and the aspect graph,'' \emph{International Journal of Computer Vision}, vol.~5, no.~2, pp. 137--160, 1990.

\bibitem{palmer1981cannonical}
S.~E. Palmer, ``Cannonical perspective and the perception of objects,'' \emph{Attention and performance}, vol.~9, pp. 135--151, 1981.

\bibitem{wang2024yolov10}
A.~Wang, H.~Chen, L.~Liu, K.~Chen, Z.~Lin, J.~Han, and G.~Ding, ``Yolov10: Real-time end-to-end object detection,'' \emph{arXiv preprint arXiv:2405.14458}, 2024.

\bibitem{zou2023object}
Z.~Zou, K.~Chen, Z.~Shi, Y.~Guo, and J.~Ye, ``Object detection in 20 years: A survey,'' \emph{Proceedings of the IEEE}, vol. 111, no.~3, pp. 257--276, 2023.

\bibitem{wu2024marvis}
J.~Wu, X.~Lin, S.~Negahdaripour, C.~Ferm{\"u}ller, and Y.~Aloimonos, ``Marvis: Motion \& geometry aware real and virtual image segmentation,'' in \emph{2024 IEEE/RSJ International Conference on Intelligent Robots and Systems (IROS)}.\hskip 1em plus 0.5em minus 0.4em\relax IEEE, 2024, pp. 2778--2785.

\bibitem{lin2024uivnav}
X.~Lin, N.~Karapetyan, K.~Joshi, T.~Liu, N.~Chopra, M.~Yu, P.~Tokekar, and Y.~Aloimonos, ``Uivnav: Underwater information-driven vision-based navigation via imitation learning,'' in \emph{2024 IEEE International Conference on Robotics and Automation (ICRA)}.\hskip 1em plus 0.5em minus 0.4em\relax IEEE, 2024, pp. 5250--5256.

\bibitem{wu2025single}
J.~Wu, T.~Wang, M.~A.~B. Siddique, M.~J. Islam, C.~Fermuller, Y.~Aloimonos, and C.~A. Metzler, ``Single-step latent diffusion for underwater image restoration,'' \emph{arXiv preprint arXiv:2507.07878}, 2025.

\bibitem{xiongevent3dgs}
T.~Xiong, J.~Wu, B.~He, C.~Fermuller, Y.~Aloimonos, H.~Huang, and C.~Metzler, ``Event3dgs: Event-based 3d gaussian splatting for high-speed robot egomotion,'' in \emph{8th Annual Conference on Robot Learning}.

\bibitem{wu20233d}
J.~Wu, B.~Yu, and M.~J. Islam, ``3d reconstruction of underwater scenes using nonlinear domain projection,'' in \emph{2023 IEEE Conference on Artificial Intelligence (CAI)}.\hskip 1em plus 0.5em minus 0.4em\relax IEEE, 2023, pp. 359--361.

\bibitem{billard2019trends}
A.~Billard and D.~Kragic, ``Trends and challenges in robot manipulation,'' \emph{Science}, vol. 364, no. 6446, p. eaat8414, 2019.

\bibitem{siddique2025aquafuse}
M.~A.~B. Siddique, J.~Wu, I.~Rekleitis, and M.~J. Islam, ``Aquafuse: Waterbody fusion for physics-guided view synthesis of underwater scenes,'' \emph{IEEE Robotics and Automation Letters}, 2025.

\bibitem{bajcsy2018revisiting}
R.~Bajcsy, Y.~Aloimonos, and J.~K. Tsotsos, ``Revisiting active perception,'' \emph{Autonomous Robots}, vol.~42, pp. 177--196, 2018.

\bibitem{aloimonos2013active}
Y.~Aloimonos, \emph{Active perception}.\hskip 1em plus 0.5em minus 0.4em\relax Psychology Press, 2013.

\bibitem{yuan2024learning}
D.~Yuan, L.~Burner, J.~Wu, M.~Liu, J.~Chen, Y.~Aloimonos, and C.~Ferm{\"u}ller, ``Learning normal flow directly from event neighborhoods,'' \emph{arXiv preprint arXiv:2412.11284}, 2024.

\bibitem{yu2022udepth}
B.~Yu, J.~Wu, and M.~J. Islam, ``Udepth: Fast monocular depth estimation for visually-guided underwater robots,'' \emph{arXiv preprint arXiv:2209.12358}, 2022.

\bibitem{wu2023low}
J.~Wu, ``Low-cost depth estimation and 3d reconstruction in scattering medium,'' Ph.D. dissertation, University of Florida, 2023.

\bibitem{oquab2023dinov2}
M.~Oquab, T.~Darcet, T.~Moutakanni, H.~Vo, M.~Szafraniec, V.~Khalidov, P.~Fernandez, D.~Haziza, F.~Massa, A.~El-Nouby \emph{et~al.}, ``Dinov2: Learning robust visual features without supervision,'' \emph{arXiv preprint arXiv:2304.07193}, 2023.

\bibitem{radford2021learning}
A.~Radford, J.~W. Kim, C.~Hallacy, A.~Ramesh, G.~Goh, S.~Agarwal, G.~Sastry, A.~Askell, P.~Mishkin, J.~Clark \emph{et~al.}, ``Learning transferable visual models from natural language supervision,'' in \emph{International conference on machine learning}.\hskip 1em plus 0.5em minus 0.4em\relax PMLR, 2021, pp. 8748--8763.

\bibitem{zhang2019overview}
Y.~Zhang and G.~Fei, ``Overview of 3d scene viewpoints evaluation method,'' \emph{Virtual Reality \& Intelligent Hardware}, vol.~1, no.~4, pp. 341--385, 2019.

\bibitem{roldao20223d}
L.~Roldao, R.~De~Charette, and A.~Verroust-Blondet, ``3d semantic scene completion: A survey,'' \emph{International Journal of Computer Vision}, vol. 130, no.~8, pp. 1978--2005, 2022.

\bibitem{palmer1981canonical}
S.~E. Palmer, E.~Rosch, and P.~Chase, ``Canonical perspective and the perception of objects,'' in \emph{International Symposium on Attention and Performance (Attention and Performance IX)}.

\bibitem{Plemenos1996}
D.~Plemenos and M.~Benayada, ``Intelligent display techniques in scene modelling: New techniques to automatically compute good views,'' in \emph{Proceedings of the International Conference on Graphics}, St. Petersburg, Russia, 1996.

\bibitem{vazquez2001viewpoint}
P.-P. V{\'a}zquez, M.~Feixas, M.~Sbert, and W.~Heidrich, ``Viewpoint selection using viewpoint entropy.'' in \emph{VMV}, vol.~1.\hskip 1em plus 0.5em minus 0.4em\relax Citeseer, 2001, pp. 273--280.

\bibitem{vieira2009learning}
T.~Vieira, A.~Bordignon, A.~Peixoto, G.~Tavares, H.~Lopes, L.~Velho, and T.~Lewiner, ``Learning good views through intelligent galleries,'' in \emph{Computer Graphics Forum}, vol.~28, no.~2.\hskip 1em plus 0.5em minus 0.4em\relax Wiley Online Library, 2009, pp. 717--726.

\bibitem{feldman2005information}
J.~Feldman and M.~Singh, ``Information along contours and object boundaries.'' \emph{Psychological review}, vol. 112, no.~1, p. 243, 2005.

\bibitem{stoev2002case}
S.~L. Stoev and W.~Stra{\ss}er, ``A case study on automatic camera placement and motion for visualizing historical data,'' in \emph{IEEE Visualization, 2002. VIS 2002.}\hskip 1em plus 0.5em minus 0.4em\relax IEEE, 2002, pp. 545--548.

\bibitem{page2003shape}
D.~L. Page, A.~F. Koschan, S.~R. Sukumar, B.~Roui-Abidi, and M.~A. Abidi, ``Shape analysis algorithm based on information theory,'' in \emph{Proceedings 2003 international conference on image processing (Cat. No. 03CH37429)}, vol.~1.\hskip 1em plus 0.5em minus 0.4em\relax IEEE, 2003, pp. I--229.

\bibitem{leifman2016surface}
G.~Leifman, E.~Shtrom, and A.~Tal, ``Surface regions of interest for viewpoint selection,'' \emph{IEEE transactions on pattern analysis and machine intelligence}, vol.~38, no.~12, pp. 2544--2556, 2016.

\bibitem{laga2010semantics}
H.~Laga, ``Semantics-driven approach for automatic selection of best views of 3d shapes,'' in \emph{3DOR@ Eurographics}, 2010, pp. 15--22.

\bibitem{attene2006mesh}
M.~Attene, S.~Katz, M.~Mortara, G.~Patan{\'e}, M.~Spagnuolo, and A.~Tal, ``Mesh segmentation-a comparative study,'' in \emph{IEEE International Conference on Shape Modeling and Applications 2006 (SMI'06)}.\hskip 1em plus 0.5em minus 0.4em\relax IEEE, 2006, pp. 7--7.

\bibitem{zhang20203d}
Y.~Zhang, G.~Fei, and G.~Yang, ``3d viewpoint estimation based on aesthetics,'' \emph{IEEE Access}, vol.~8, pp. 108\,602--108\,621, 2020.

\bibitem{gualtieri2017viewpoint}
M.~Gualtieri and R.~Platt, ``Viewpoint selection for grasp detection,'' in \emph{2017 IEEE/RSJ International Conference on Intelligent Robots and Systems (IROS)}.\hskip 1em plus 0.5em minus 0.4em\relax IEEE, 2017, pp. 258--264.

\bibitem{Kiciroglu_2020_CVPR}
S.~Kiciroglu, H.~Rhodin, S.~N. Sinha, M.~Salzmann, and P.~Fua, ``Activemocap: Optimized viewpoint selection for active human motion capture,'' in \emph{Proceedings of the IEEE/CVF Conference on Computer Vision and Pattern Recognition (CVPR)}, June 2020.

\bibitem{chen2024active}
J.~Chen, B.~He, C.~D. Singh, C.~Fermuller, and Y.~Aloimonos, ``Active human pose estimation via an autonomous uav agent,'' \emph{arXiv preprint arXiv:2407.01811}, 2024.

\bibitem{song2022unsupervised}
R.~Song, W.~Zhang, Y.~Zhao, and Y.~Liu, ``Unsupervised multi-view cnn for salient view selection and 3d interest point detection,'' \emph{International Journal of Computer Vision}, vol. 130, no.~5, pp. 1210--1227, 2022.

\bibitem{siming2024active}
H.~Siming, C.~D. Hsu, D.~Ong, Y.~S. Shao, and P.~Chaudhari, ``Active perception using neural radiance fields,'' in \emph{2024 American Control Conference (ACC)}.\hskip 1em plus 0.5em minus 0.4em\relax IEEE, 2024, pp. 4353--4358.

\bibitem{mildenhall2020nerfrepresentingscenesneural}
B.~Mildenhall, P.~P. Srinivasan, M.~Tancik, J.~T. Barron, R.~Ramamoorthi, and R.~Ng, ``Nerf: Representing scenes as neural radiance fields for view synthesis,'' 2020.

\bibitem{Blanz1999WhatOA}
V.~Blanz, M.~J. Tarr, and H.~H. B{\"u}lthoff, ``What object attributes determine canonical views?'' \emph{Perception}, vol.~28, pp. 575 -- 599, 1999.

\bibitem{deitke2022objaverseuniverseannotated3d}
M.~Deitke, D.~Schwenk, J.~Salvador, L.~Weihs, O.~Michel, E.~VanderBilt, L.~Schmidt, K.~Ehsani, A.~Kembhavi, and A.~Farhadi, ``Objaverse: A universe of annotated 3d objects,'' 2022.

\bibitem{deng2009imagenet}
J.~Deng, W.~Dong, R.~Socher, L.-J. Li, K.~Li, and L.~Fei-Fei, ``Imagenet: A large-scale hierarchical image database,'' in \emph{2009 IEEE conference on computer vision and pattern recognition}.\hskip 1em plus 0.5em minus 0.4em\relax Ieee, 2009, pp. 248--255.

\bibitem{blender}
B.~O. Community, \emph{Blender - a 3D modelling and rendering package}, Blender Foundation, Stichting Blender Foundation, Amsterdam, 2018.

\bibitem{loshchilov2019decoupledweightdecayregularization}
I.~Loshchilov and F.~Hutter, ``Decoupled weight decay regularization,'' 2019.

\bibitem{Unity2024}
{Unity Technologies}, ``Unity,'' 2024.

\bibitem{quigley2009ros}
M.~Quigley, K.~Conley, B.~Gerkey, J.~Faust, T.~Foote, J.~Leibs, R.~Wheeler, A.~Y. Ng \emph{et~al.}, ``Ros: an open-source robot operating system,'' in \emph{ICRA workshop on open source software}, vol.~3, no. 3.2.\hskip 1em plus 0.5em minus 0.4em\relax Kobe, Japan, 2009, p.~5.

\bibitem{kam2015rviz}
H.~R. Kam, S.-H. Lee, T.~Park, and C.-H. Kim, ``Rviz: a toolkit for real domain data visualization,'' \emph{Telecommunication Systems}, vol.~60, pp. 337--345, 2015.

\end{thebibliography}

\end{document}